\documentclass[letterpaper, 10 pt, conference]{ieeeconf}  

\IEEEoverridecommandlockouts                              
\usepackage{graphicx} 
\usepackage{float}
\usepackage{subfig}
\newcommand{\etal}{\textit{et al}.}

\title{\LARGE \bf Efficient LiDAR data compression for embedded V2I or V2V data handling}

\author{Paul CAILLET$^{1}$ and Yohan DUPUIS$^{2}$
\thanks{*This work is a part of COPTER project, which has been selected in the context of Normandy Major Research Network (GRR) co-financed by ERDF and Normandy Region}
\thanks{$^{1}$Paul CAILLET is with Department of Multimodal Transportation Infrastructure, CEREMA, 76120 Le Grand Quevilly, France {\tt\small paul.caillet@cerema.fr}}
\thanks{$^{2}$Yohan DUPUIS is with IRSEEM, ESIGELEC, 76800 Saint-Etienne-du-Rouvray, France {\tt\small yohan.dupuis@esigelec.fr}}}

\begin{document}

\maketitle
\thispagestyle{empty}
\pagestyle{empty}

\begin{abstract}
LiDAR are  increasingly being used in intelligent vehicles (IV) or intelligent transportation systems (ITS). Storage and transmission of data generated by LiDAR sensors are one of the most challenging aspects of their deployment. In this paper we present a method that can be used to efficiently compress LiDAR data in order to facilitate storage and transmission in V2V or V2I applications. This method can be used to perform lossless or lossy compression and is specifically designed for embedded applications with low processing power. This method is also designed to be easily applicable to existing processing chains by keeping the structure of the data stream intact. We benchmarked our method using several publicly available datasets and compared it with state-of-the-art LiDAR data compression methods from the literature.
\end{abstract}

\begin{keywords}
LiDAR, embedded systems, roadside measurement, compression.

\end{keywords}

\section{INTRODUCTION}

Light Detection And Ranging (LiDAR) systems are effective at mapping their environment quickly and accurately. Their are robust to various factors that may challenge computer vision based systems. These devices are being used more and more in various fields such as autonomous vehicles or roadside measurement units.

IV and ITS success relies on being able to transmit and share data from sensors that may augment the vehicle or infrastructure information among vehicle-to-vehicle (V2V), vehicle-to-infrastructure (V2I) frameworks to improve road safety and operation. Computing power on roadside measurement units are highly optimized as they require batteries to deal with power supply instability. Consequently, the processing power required to interpret complex data is often deported into the Traffic Monitoring Centre. 

One of the biggest challenges in the operation of such devices is the storage or transmission of the generated data stream. Modern 3D LiDAR systems are able to generate up to several millions of points of measure per second. Such a data stream requires bandwidths that exceed the capabilities of most modern wireless network technologies.

Severals methods had been proposed to tackle LiDAR data storage and transmission on embedded platforms that stands real-time processing. 

The American Society of Photogrammetry and Remote Sensing (ASPRS) created a standardized binary exchange format named LAS for practical exchange of LiDAR data without the limitations of vendor-specific binary formats \cite{c1}. 
Isenburg presented in 2012 LASzip, a lossless, non-progressive data compressor for the LAS format \cite{c2} which allow for dynamic decompression of specific parts of the dataset.

For lossy compression, an intermediate representation of the point cloud is often used. Tree-like data structure and depth maps are the most commonly used representations. 
Meagher presented in 1982 a way to encode arbitrary 3-D objects with an arbitrary resolution using an octree \cite{c3}. Hornung~\etal~\cite{c5} used an octree to develop a probabilistic mapping framework, which allow for live mapping of the environment and constant updating of its probabilistic model.
Houshiar ~\etal~\cite{c6} used conventional image compression techniques on panorama images generated from spherical coordinates of measured points. This approach requires converting LiDAR raw data to images as well as storing azimuths information in separate images when using Velodyne LiDARs.
Nenci~\etal~\cite{c7} used the H.264 video encoder to compress range data streams, allowing for live transmission of the data. Range measurements are split in separate channels and re-sampled with shorter integer values. A video encoding thread is ran for each range channel and the decompression is done by merging and re-sampling every channel into one.

In this paper, we address the problem of fast compression of raw LiDAR data for storage and real time transmission on embedded devices. Our solution relies on the binary representation of the data generated by most LiDAR systems and the unnecessary precision in the encoding of the measurements. Bit masking is used to quickly re-sample measurements to a lower number of bits by zeroing least significant bits. The number of zeroed bits can be tuned to retain sufficient accuracy while allowing for significant compression ratios thanks to the redundant patterns generated in the data.

Our method is designed to be used in embedded applications with low available processing power. By quickly compressing the data it allows for real-time data transmission on low bandwidth networks. This method has also been designed to be fully compatible with existing processing chains by keeping the original structure of the data data stream.

This paper is organized as follows: Section~\ref{sec:methodo} presents our method. Section~\ref{sec:results} and Section~\ref{sec:comparison} develops the results we obtained. We conclude and discuss the results in Section~\ref{sec:conclusion}.

\section{Methodology}
\label{sec:methodo}

\subsection{Data compression using bit masking}

Our method operate on the raw LiDAR data stream as generated by LiDAR sensors. A bit mask is applied on the raw data packet in order to set to zero the $n$ least significant bits (LSB) of each measurement, thus creating repeating zero patterns.
A conventional lossless data compression algorithm is then used to compress the raw data. This method preserves the structure of the raw data packets, making it very easy to integrate in an existing tool chain.

\subsection{Measurements representation}
Most of the time, LiDAR range measurements are represented using unsigned integers. Absolute distance is obtained by multiplying the given integer by a device-specific step size (e.g. 2 mm). In most cases, this step size is largely inferior to the accuracy rating given by the vendor (e.g. 2 mm / $\pm$3 cm for the Velodyne VLP-16). This gap between the measurement representation and the accuracy of the device can be exploited to reduce the size of the data while keeping the measurements within the accuracy rating given by the device's vendor.

\subsection{Bit masking}
\begin{figure}[t]
    \centering
    \includegraphics[width=\linewidth]{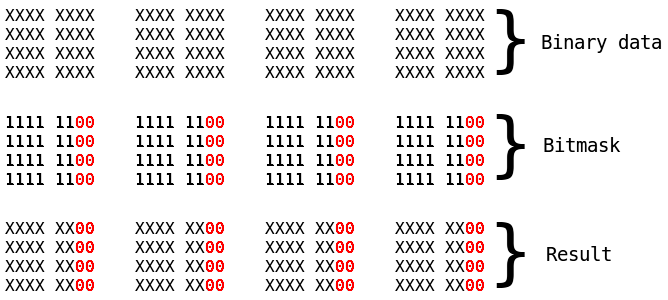}
    \caption{Bit masking operation, here with 8 bits integer and 2 bits masked. Note the recurrent patterns of zeros in the resulting sequence}
    \label{fig:bitmask}
\end{figure}

A binary mask is used to set the $n$ least significant bits of each measurement to zero (Figure~\ref{fig:bitmask}). Each bit of the mask is set to one except the bits covering the $n$ least significant bits of the measurements. The mask is then applied to the data using a logical AND operation, thus conserving the state of the bits masked by a one and putting the bits masked by a zero to zero.

The masking operation can be performed in several other ways: right shifts followed by left shifts or divisions by two followed by left shifts. These operations require several CPU cycles to be performed and makes batch processing more complicated to implement.
\newline
Bit masking has the advantage of being much faster: on most architectures a logical AND is performed in one CPU cycle and we can process the data in bulk using a large bit mask. In our implementation, a bit mask covering the whole data packet is stored in memory and applied to each incoming data packet.

\subsection{Loss of accuracy}
The loss of accuracy caused by our method can be precisely qualified. The bit masked distance $d'$ is always smaller or equal to the original distance $d$.

\begin{equation}
    d' \leq d
    \label{eq:1}
\end{equation}

Moreover, depending on the number of bits masked $n$, it is possible to bind the error $err$ on each measurement $d$, knowing the device-specific step size $s$:

\begin{equation}
    0 \leq | err | < 2^n*s
    \label{eq:2}
\end{equation}

The maximum error increases exponentially with the number of bits masked, but the error remains acceptable with $n$ being reasonably small ($n \leq 6$ for 16 bits measurements).
\begin{figure}[t]
    \centering
    \subfloat[Road side unit]{\includegraphics[width=0.9\linewidth]{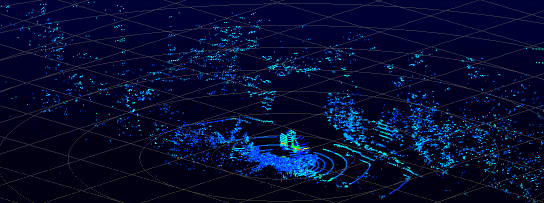}\label{piste}}\\
    \subfloat[Onboard unit]{\includegraphics[width=0.9\linewidth]{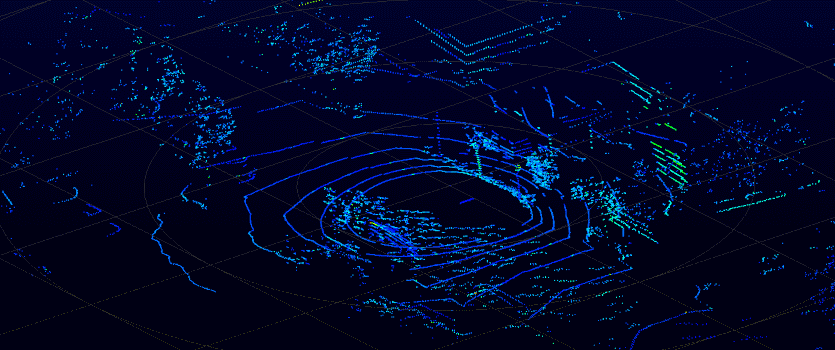}\label{velodyne}}
    \caption{Experiment dataset overviews}
\end{figure}
\subsection{Data compression}
Once the raw data packets have been bit masked, they are compressed using conventional lossless data compression algorithms. This method is not bound to any data compression algorithm and one can use the algorithm that fits their needs the best, being computational power limits, storage space or low latency.
Most of the available compression algorithms can exploit the patterns of zeros produced by the bit masking and achieve higher compression ratio. In our experiment detailed later in this paper, we compared four compression algorithms to demonstrate the efficiency of our method regardless of the compression algorithm used.

\section{Experimental setup}
\label{sec:setup}
The primary goal of our experiment is to demonstrate the efficiency of our method at compressing a raw LiDAR data stream while preserving an acceptable accuracy. In our experiment, we first evaluate the evolution of the loss of accuracy with different number of zeroed bits. Then, we measure the gains induced by this operation after the data has been compressed using several widespread, general purpose, lossless data compression algorithms. Finally we evaluate the running time of the whole compression process on a desktop PC as well as several SoC (System on Chip) boards to demonstrate the viability of our method for real-time embedded applications.

\subsection{Datasets}
For our evaluation, we used three datasets acquired using a Velodyne VLP-16 range scanner running at 10Hz. These datasets present very different features and will demonstrate the robustness of our method.

The first dataset is a still recording of our office space, with mostly short distances, flat surfaces (walls) and with some distant patches visible through the windows. The second dataset (Figure \ref{piste}) is a still recording on the side of the road at our test track\footnote{Data available upon request}. Several vehicles passes by the sensor and most of the scene is noise from the forest in the background. The last dataset (Figure \ref{velodyne}) is a mobile recording of an urban area taken from a car. This data set is from the sample data provided by Velodyne with the VLP-16 sensor and publicly available\footnote{Available at https://velodynelidar.com/downloads.html}.
The first dataset has a duration of 53s, 103s for the second dataset and 133s for the third dataset.

\subsection{Compression algorithms}
We used several widespread, general-purpose lossless compression algorithms to outline the viability of our method independently of the compression algorithm used. We tested GZIP \cite{c8}, BZIP2 \cite{c9}, LZMA \cite{c10} and LZ4 \cite{c11} each with the implementation available on public Linux repositories. These compression algorithms offer several compression presets, allowing the user to adjust the compression level and computing time. We have not used these settings and kept them to their default values.

\subsection{Test machines}
The running time evaluation has been performed on a desktop computer equipped with an Intel Xeon E5-2620 CPU (6-core, 2.0GHz)  as well as on two SoC boards: a Raspberry Pi 3 and an Odroid XU, both using quad-core ARM CPUs. We ran the tests either using only one thread or using all the available CPU cores. This way, we can precisely evaluate the computational workload as well as the viability of our method in real world embedded applications. All data have been stored in RAM when possible to eliminate any latency due to disk I/O and the full benchmark was run 3 times to average the running times.

The desktop computer and the Odroid are running Ubuntu 16.04, the Raspberry Pi is running Raspbian Stretch. For compression, we used the implementations available from each distribution's package repository. The bit-masking program was written in C, compiled with GCC on each computer with the -O3 flag and no further optimization.


\section{Experimental results}
\label{sec:results}

\subsection{Accuracy of the compressed measurements}
\begin{figure}[htbp]
    \centering
    \includegraphics[width=0.8\linewidth]{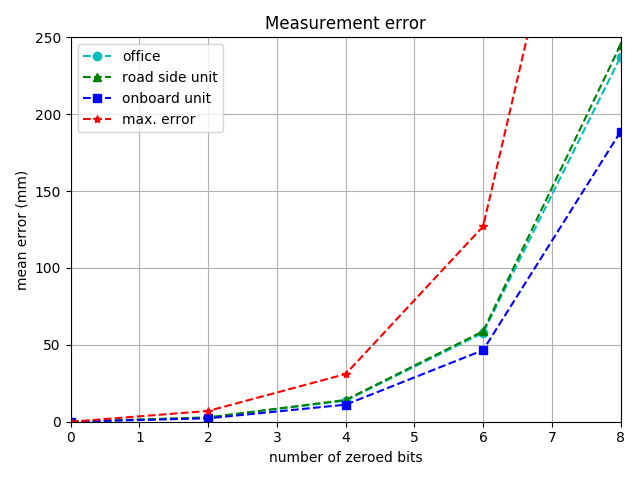}
    \includegraphics[width=0.8\linewidth]{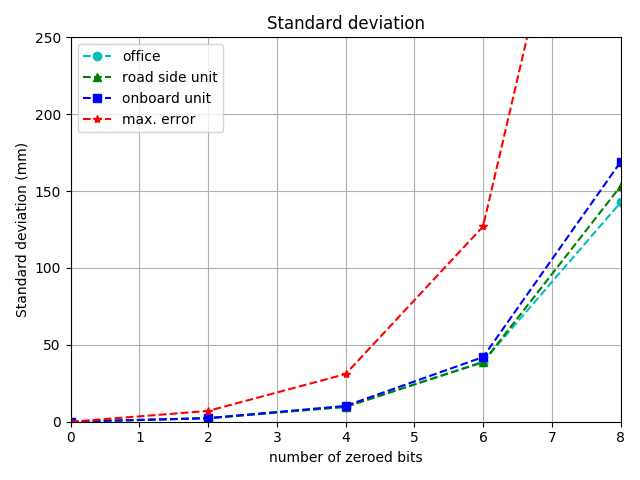}
    \caption{Evolution of the error according to the number of masked bits}
    \label{error}
\end{figure}

For our evaluation, we define the error of a measurement as the absolute difference between the original and the modified measurement. For each dataset, we computed the mean error as well as the standard deviation of the error. Figure \ref{error} also shows the maximum error, represented by the red line. The maximum error for $n = 8$ bits (510 mm) is not shown on the graph as it would make the graph less readable around meaningful values.

With $n \leq 4$ masked bits the measured and maximum error remains within the confidence interval given by the vendor ($\pm$3 cm). With $n > 4$ masked bits the measured error is greater than the confidence interval and the retained accuracy may not fit all use cases. It is interesting to note the similarity between the mean error and the standard deviation, with the standard deviation being slightly smaller than the mean error. This behavior can be explained by the systematic under-evaluation of the measurements described with equation (\ref{eq:1}).

\subsection{Compressed data size}
In this section of our experiments, we measured the efficiency of our method at compressing raw capture files. We ran each compression algorithms on the raw capture files as well as on the masked ones to get a reference point for each dataset.

\begin{figure}[htbp]
    \centering
    \includegraphics[width=0.8\linewidth]{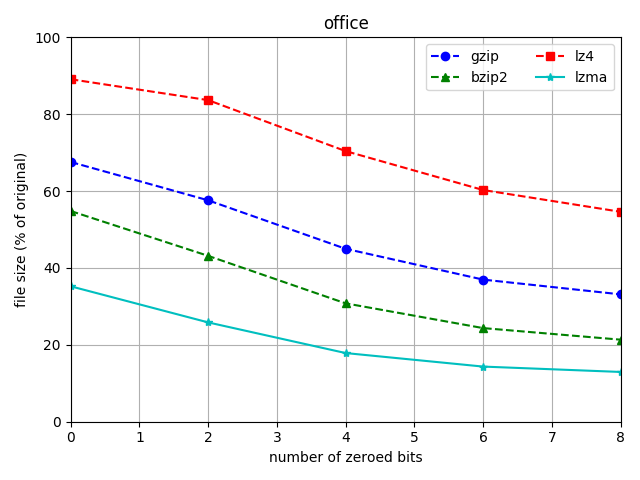}
    \includegraphics[width=0.8\linewidth]{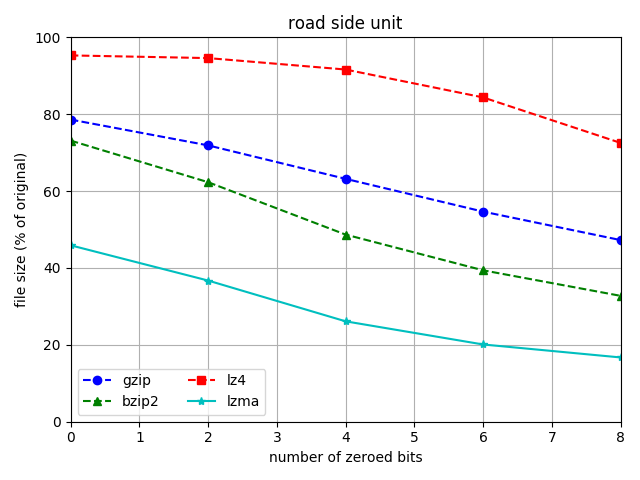}
    \includegraphics[width=0.8\linewidth]{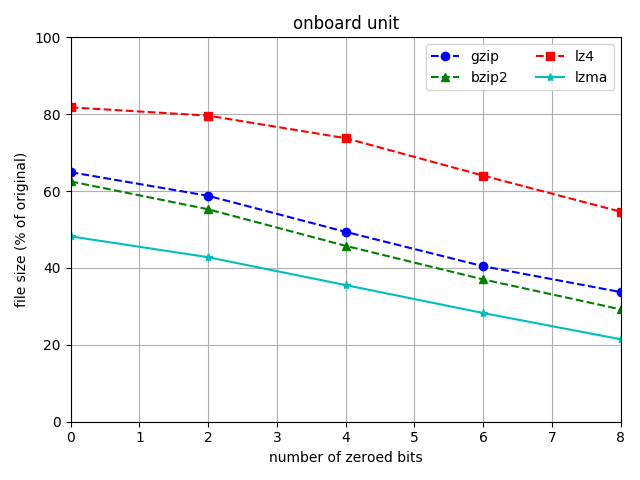}
    \caption{Relative file size according to the number of zeroed bits}
    \label{size}
\end{figure}

Without modifying the capture files, we can see that lossless compression algorithms are quite capable at compressing raw LiDAR data, with all datasets being reduced to at most 50\% of their original size after being compressed by the LZMA algorithm (Figure \ref{size}). The file sizes decreases almost linearly as more bits are masked. This evolution is similar for each dataset independently of the compression algorithm used. Absolute compression ratios are very heterogeneous and are ultimately tied to the dataset. The test track dataset which is very noisy presents low compression ratios up to the point ($ n = 4 $ bits masked) where the noise becomes smaller than the step size. In contrast, the office dataset present a high compression ratio with no bits masked but the gains are marginally smaller.
\newline
Overall, bit masking proves to be effective and enables compression algorithms to reach much higher compression ratios, but the absolute attainable compression ratio depends on the dataset and the compression algorithm used.

\subsection{Running time}
\begin{figure}[t]
    \centering
    \includegraphics[width=0.8\linewidth]{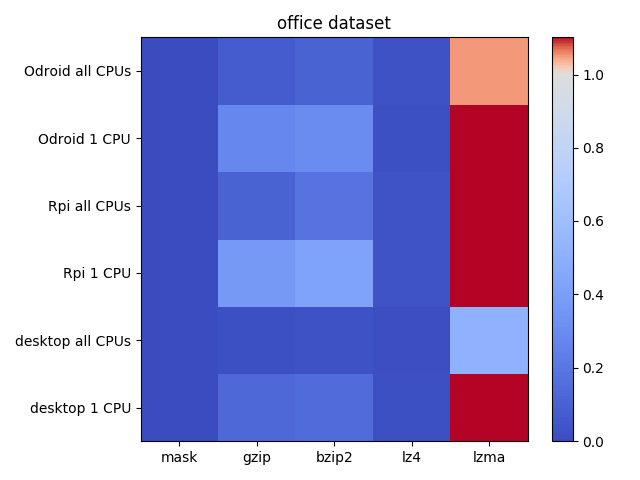}
    \includegraphics[width=0.8\linewidth]{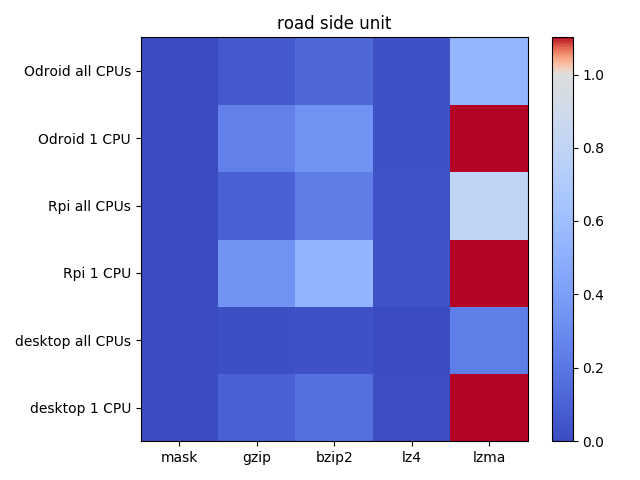}
    \includegraphics[width=0.8\linewidth]{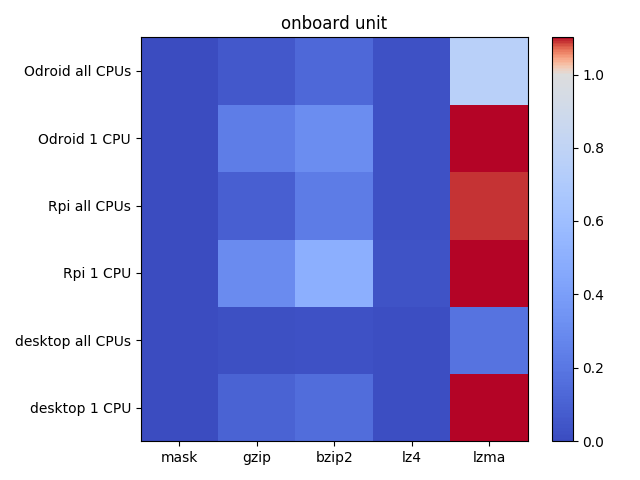}
    \caption{Normalized running time on several architectures, using one or all available CPUs. A normalized time inferior to one (blue) means the data can be processed in real-time}
    \label{tps}
\end{figure}

\begin{figure*}[ht]
    \centering
    \subfloat[Measurement Error]{\includegraphics[width=0.4\linewidth]{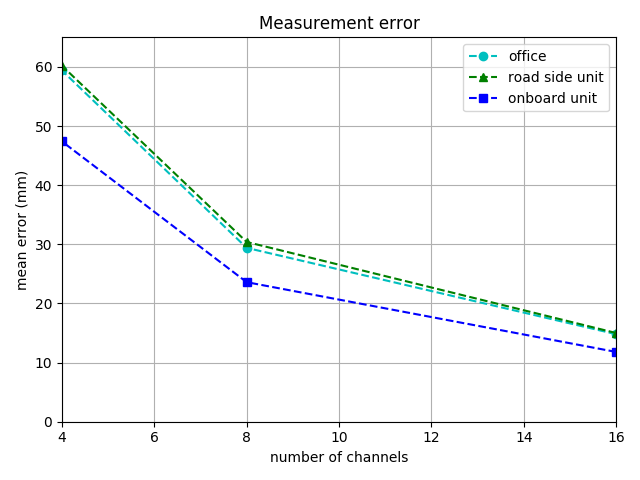}}~
    \subfloat[Standard deviation]{\includegraphics[width=0.4\linewidth]{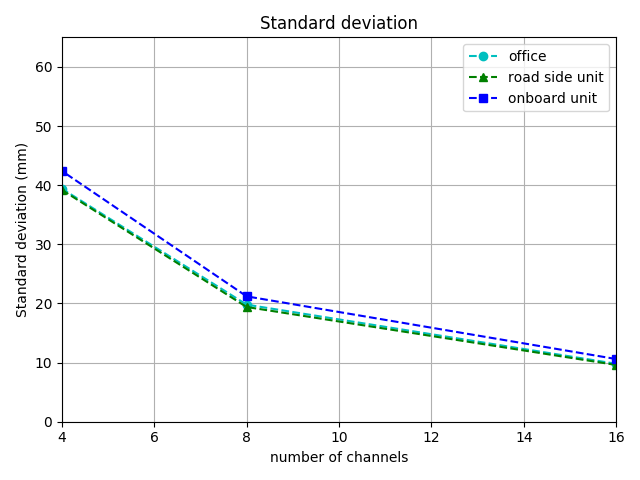}}\\
    \subfloat[Relative file size]{\includegraphics[width=0.4\linewidth]{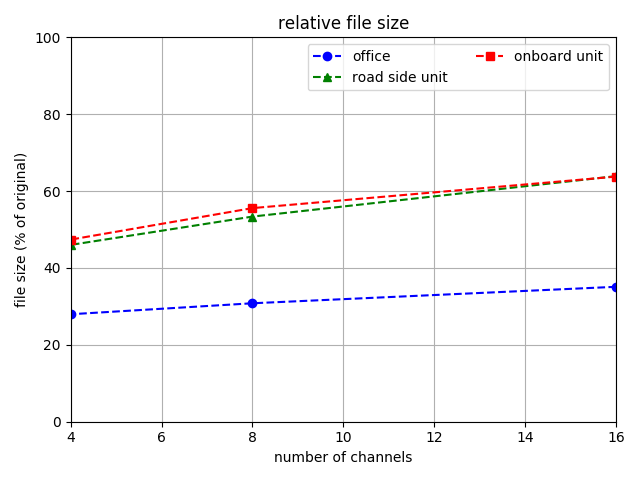}}~
    \subfloat[Relative processing time]{\includegraphics[width=0.4\linewidth]{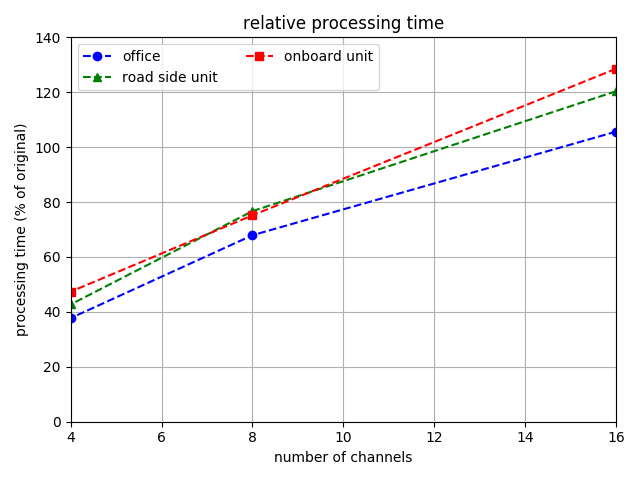}}
    \caption{H.264 performance. The video compression is lossless ($QP = 0$)}
    \label{err-h264}
\end{figure*}

Figure \ref{tps} shows the running time of the compression chain relative to the duration of the recording. The first column represents the time taken for the bit masking operation alone, while the other columns represents the time taken by the bit masking operation plus the compression using each column's respective algorithm. A value inferior to one (in blue) means that the time taken by the bit masking operation plus the data compression is inferior to the duration of the dataset, making it possible to store or transmit compressed data in real time without data loss.
The masking operation takes on average $3.8 * 10^{-6}$ seconds per packet, with the LiDAR producing a packet every $1.33 * 10^{-3}$ seconds. Most of the time is effectively spent compressing the data with the bit masking taking a negligible amount of time.

Overall, the same pattern appear across all the datasets, with every tested computer being able to run the compression in real time, except the LZMA algorithm, which is much slower.
It is important to note that compression time can fluctuate significantly depending on the data. The compression time with the LZMA algorithm is a great illustration: using all CPUs, it is possible to compress the test track data set in real time on every tested computer, while it is not possible with the office dataset.

\begin{figure*}[htb]
    \centering
    \subfloat[Original]{\includegraphics[width=0.3\linewidth]{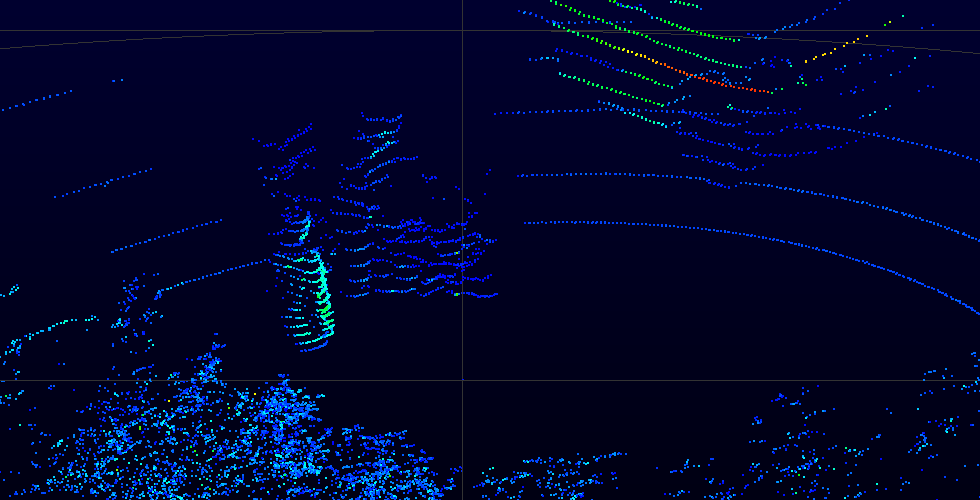}\label{}}~
    \subfloat[n = 4 bits]{\includegraphics[width=0.3\linewidth]{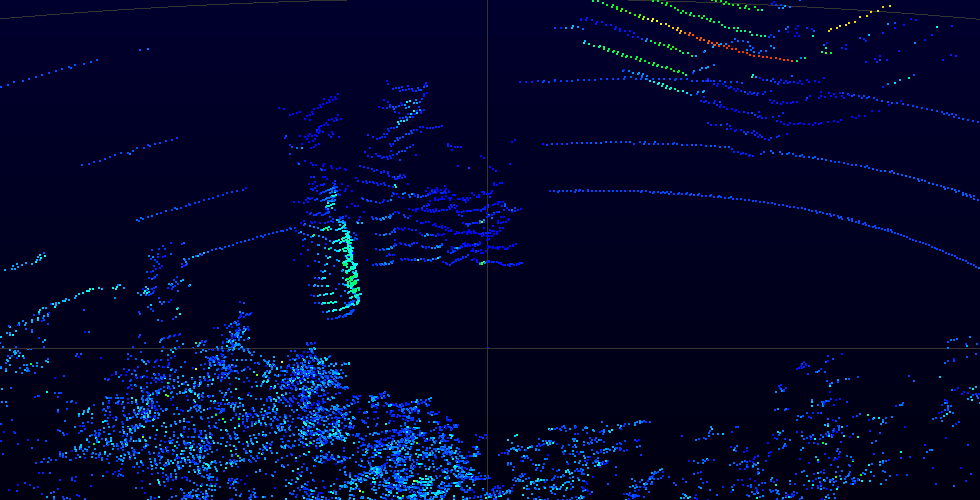}\label{}}~
    \subfloat[n = 8 bits]{\includegraphics[width=0.3\linewidth]{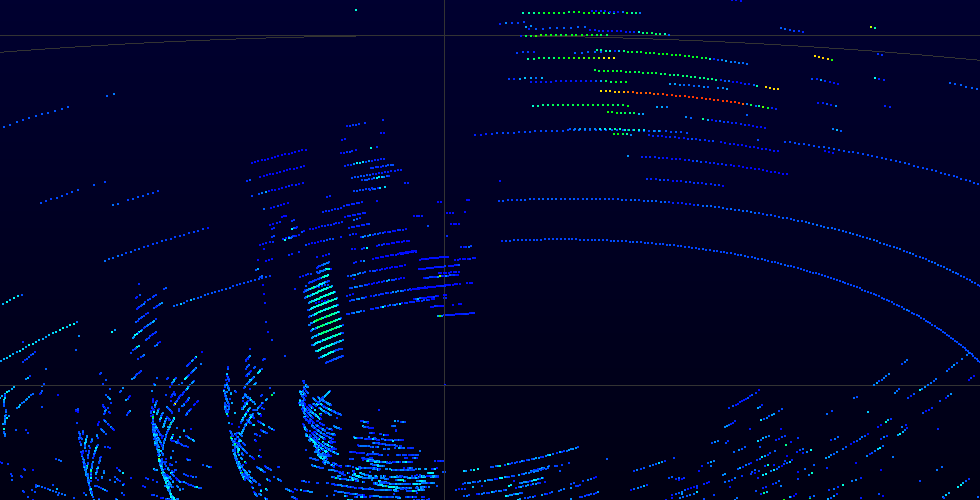}\label{}}
    \caption{Sample view for different quality presets}
    \label{result}
\end{figure*}

\section{Comparison with other methods}
\label{sec:comparison}
We compared our method to some of the methods presented in section I. Some solutions were quickly pushed away as they did not fit our goal of real-time processing on embedded hardware.\\
For the comparison, we defined two indicators: RPT = processing time / dataset duration and RFS = compressed file size / dataset size
\subsection{Octomap}
We used the ROS (Robot Operating System \cite{c12}) distribution of the Octomap framework for our tests. We could not achieve real-time processing of our datasets while keeping a decent resolution (grid size of 10cm) even on desktop hardware.

\subsection{PNG compression}
We generated a series of panorama images from the LiDAR data and compressed it with the PNG algorithm. The compression is lossless, all distance measurements are kept with their original 16 bits resolution. The intensity is encoded on another set of 8 bits gray images. The running time has been measured on a desktop computer, constrained to a single thread. For comparison, we included in Table \ref{table_size} and Table \ref{table_time} the result obtained using BZIP2 on the raw capture files.

\begin{table}[htbp]
    \begin{center}
        \begin{tabular}{|l|c|c|}
      \hline
      Dataset & PNG & BZIP2 \\
      \hline
      office & 47.5 \% & 55.1 \% \\
      test track & 64.2 \% & 72.6 \% \\
      Velodyne sample & 48.3 \% & 63.1 \% \\
      \hline
    \end{tabular}
    \end{center}
    \caption{Relative file size}
    \label{table_size}
\end{table}

\begin{table}[htbp]
    \centering
    \begin{tabular}{|l|c|c|}
      \hline
      Dataset & PNG & BZIP2 \\
      \hline
      office & 24.1 \% & 15.2 \% \\
      test track & 21.0 \% & 17.4 \% \\
      Velodyne sample & 19.4 \% & 15.2 \% \\
      \hline
    \end{tabular}
    \caption{Relative processing time}
    \label{table_time}
\end{table}

The PNG image format is efficient at compressing LiDAR data, with performances comparable to a lossless compression algorithm. We believe it is a good alternative to our method for lossless compression of LiDAR data. However, PNG being a lossless file format, it cannot achieve compression ratios as high as other lossy methods.

\subsection{Video compression}
We benchmarked the method presented in \cite{c7} using the implementation provided by the authors on the three datasets presented previously, with the process being bound to a single CPU on the desktop computer. The video compression is always lossless ($QP = 0$) (Figure~\ref{err-h264}).

\begin{table}
\centering
\subfloat[Our approach (4 zeroed bits + bzip2)]{\begin{tabular}{|c|c|c|c|c|}
	\hline 
	Dataset& $\mu$ (mm)& $\sigma$ (mm)  & RFS ($\%$) & RPT ($\%$)\\ 
	\hline 
	Office& 13.6 & 9.7 & 30.7 & 14.7\\ 
	\hline 
	Test track& 14.1 & 9.6 & 48.6 & 16.9 \\ 
	\hline 
	Velodyne Sample& 11.0 & 10.3 & 45.7 & 15.3 \\ 
	\hline 
\end{tabular}}

\subfloat[H.264~\cite{c7} with 16 channels]{\begin{tabular}{|c|c|c|c|c|}
	\hline 
	Dataset& $\mu$ (mm)& $\sigma$  (mm)& RFS ($\%$) & RPT ($\%$)\\ 
	\hline 
	Office& 14.8 & 9.8 & 35.1 & 105\\ 
	\hline 
	Test track& 15.0 & 9.6 & 63.8 & 120\\ 
	\hline 
	Velodyne Sample& 11.8 & 10.6 & 63.7 & 128\\ 
	\hline 
\end{tabular}}
\caption{Performance Comparison with state-of-the art method}
\label{tab:bench}
\end{table}

The evolution of the error with the video compression method is similar to the one obtained with our method (\ref{error}). Doubling the number of channels roughly divide the error and standard deviation by two.
The file size shows a similar evolution, with the file size decreasing as the number of channels decreases.
The biggest difference comes from the processing time. Adding channels means more H.264 streams to encode, which increases the processing time.

The approach presented in ~\cite{c7} can be compared to our approach when the number of channels is a power of two. In fact, $2^c$ channels of 256 bits is able to store $2^{c+8}$ distinctive values. Our approach actually stores $2^{16-n}$ distinctive values, with $n$ the number of zeroed LSB. Table~\ref{tab:bench} compares the performance  for $c=4$ and $n=4$.

Our method present better results for each criteria. By choosing a number of zeroed bits and a number of channels producing a similar error, compression ratio and the processing times are better using our method. The processing time difference between both methods is significant. Our tests showed that running the video compression method on embedded hardware is roughly 2 to 3 times slower than on a desktop computer, which makes this method not suited for embedded applications.

\section{CONCLUSIONS}
\label{sec:conclusion}
We proposed in this paper an efficient method for compressing raw LiDAR data streams in embedded applications. This method can be used for real-time transmission, greatly reducing the bandwidth required for the data transfer, or for data archiving, reducing the storage space required. It is possible to perform lossy compression, allowing for higher compression ratios with a fine control on the error induced by the alteration of measurements. We performed a series of test on several heterogeneous datasets and different types of computer to demonstrate the efficiency of our method, especially in real-world conditions. We also compared our method to several other methods to further demonstrate its efficiency. Future works include benchmarks on other LiDARs. 





%


\end{document}